\RequirePackage{color}
\definecolor{subsectioncolor}{rgb}{0,0,0}
\documentclass[journal,twoside,web]{ieeecolor}
\providecommand{\refname}{References}
\usepackage{generic}
\usepackage{cite}
\usepackage{amsmath,amssymb,amsfonts}
\usepackage{algorithmic}
\usepackage[final]{graphicx}
\usepackage{textcomp}
\usepackage{xcolor}
\usepackage{booktabs}
\usepackage{array}
\usepackage{multirow}
\usepackage{stfloats}
\usepackage{subfig}
\usepackage{url}
\usepackage{epstopdf}
\usepackage{titlesec}
\titlespacing*{\section}{0pt}{1.0ex plus 0.2ex minus 0.2ex}{0.6ex}
\titlespacing*{\subsection}{0pt}{0.8ex plus 0.2ex minus 0.2ex}{0.4ex}

\graphicspath{{figures/}{./}}

\begin{document}

\title{Uncertainty-Aware Prediction of Lung Tumor Growth from Sparse Longitudinal CT Data via Bayesian Physics-Informed Neural Networks}

\author{
Lingfei Kong,
Haoran Ma
\thanks{
Lingfei Kong is with the Department of Mathematics, Vanderbilt University, Nashville, TN, USA (e-mail: lingfei.kong@vanderbilt.edu).
}
\thanks{
Haoran Ma is with the John A. Paulson School of Engineering and Applied Science,
Harvard University, Allston, MA, USA
(e-mail: haoran\_ma@g.harvard.edu).
Corresponding author.
}
}

\maketitle
\markboth{Kong and Ma: Uncertainty-Aware Prediction of Lung Tumor Growth via Bayesian Physics-Informed Neural Networks}{}
\thispagestyle{empty}
\pagestyle{empty}

\begin{abstract}
This work studies lung tumor growth prediction from sparse and irregular longitudinal computed tomography (CT) observations with measurement variability. A Bayesian physics-informed neural network is developed by combining Gompertz growth dynamics with low-dimensional Bayesian inference in the log-volume domain. The framework employs a two-stage inference strategy combining maximum a posteriori (MAP) estimation and Hamiltonian Monte Carlo (HMC) sampling to estimate posterior predictive distributions and uncertainty intervals. The method was evaluated on longitudinal data from the National Lung Screening Trial (30 patients). Results show that the model captures heterogeneous tumor growth patterns while maintaining reasonable prediction accuracy under limited observations. Compared with deterministic modeling approaches, the proposed approach additionally provides calibrated uncertainty estimates. The inferred posterior parameter correlations were consistent with expected biological growth behavior. The proposed framework achieved a cohort-level log-space RMSE of approximately 0.20 together with well-calibrated 95\% credible interval coverage across 30 patients. These findings suggest that Bayesian physics-informed modeling may be useful for uncertainty-aware tumor growth assessment when only limited longitudinal follow-up scans are available.
\end{abstract}


\section{Introduction}
\IEEEPARstart{L}{ung} cancer remains one of the leading causes of cancer-related morbidity and mortality worldwide. Early screening and longitudinal follow-up are important components of lung cancer management and treatment planning. The widespread adoption of low-dose computed tomography (CT) screening---supported by the NLST trial, which demonstrated a reduction in lung cancer-specific mortality~\cite{nlst2011}---has increased the availability of longitudinal imaging data in clinical practice. However, such observations are typically sparse, irregularly sampled, and affected by measurement variability, rendering reliable predictions and uncertainty quantification challenging. In clinical decision-making, predictive uncertainty is often as important as point estimates because it may influence follow-up planning and clinical risk assessment~\cite{ocana2024}.

Classical tumor growth models are typically based on mechanistic kinetic formulations, including the Gompertz model~\cite{swanson2001,byrne1995,norton1988}. These models provide biologically interpretable descriptions of tumor dynamics using a small number of parameters~\cite{lorenzo2024,norton1988}; however, they are commonly estimated through deterministic optimization procedures and therefore have limited capability for characterizing observational uncertainty and inter-patient variability. In contrast, data-driven approaches, such as Gaussian processes and deep neural networks~\cite{deisenroth2015,nickisch2008}, are effective at modeling complex nonlinear relationships directly from data. Nevertheless, these approaches generally lack explicit biological constraints, and their stability and interpretability may deteriorate in small-sample or extrapolation settings, particularly in parameter identification problems~\cite{liu2025}.

Physics-Informed Neural Networks (PINNs) provide a mechanism for integrating data-driven learning with governing physical equations by incorporating equation residuals into the training objective. Prior studies have shown that PINNs can improve generalization and robustness under limited-data conditions~\cite{raissi2019,wang2021a,wang2022,mishra2022}. Recent studies have further explored uncertainty-aware and physics-constrained learning frameworks for tumor growth modeling and related biomedical applications~\cite{yang2021,costa2024,zhu2018,rodrigues2024,kisbenedek2026}. Related scientific machine learning frameworks, such as Universal Differential Equations (UDEs), further integrate mechanistic dynamics with neural network representations~\cite{rackauckas2020}. In parallel, approximate Bayesian deep learning approaches, including Bayesian neural networks and dropout-based variational approximations~\cite{neal1996,gal2016}, have also been investigated for scalable uncertainty estimation. However, these approaches differ substantially from the explicit Hamiltonian Monte Carlo (HMC) posterior inference adopted in this work.

Despite these developments, uncertainty-aware modeling of longitudinal CT follow-up data remains challenging because clinical observations are often sparse and temporally irregular. Existing approaches typically prioritize either mechanistic consistency or uncertainty-aware learning, but rarely address both simultaneously under sparse longitudinal clinical observations. In particular, uncertainty quantification for physics-constrained tumor growth modeling remains difficult when only a few irregular follow-up scans are available per patient~\cite{betancourt2017,duane1987,chen2014,hoffman2014}. As a result, developing biologically constrained prediction models with reliable uncertainty estimation remains an active research challenge in longitudinal tumor growth modeling.

\begin{figure}[!t]
\centering
\includegraphics[width=\linewidth]{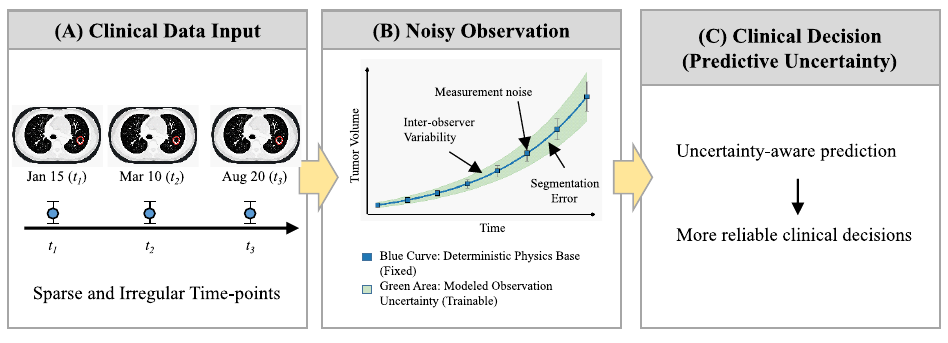}
\caption{Overview of the Bayesian PINN framework for longitudinal tumor growth prediction. Sparse and irregular CT observations are used to estimate predictive trajectories and uncertainty intervals under Gompertz-constrained posterior inference. The shaded region represents predictive uncertainty.}
\label{fig:overview}
\end{figure}

The contributions of this study can be summarized as follows:
\begin{enumerate}
\item A Bayesian physics-informed framework is presented for uncertainty-aware tumor growth prediction under sparse longitudinal CT follow-up, combining mechanistic Gompertz constraints with probabilistic inference in a clinically realistic low-data setting.
\item To improve computational feasibility and parameter interpretability, posterior inference is restricted to a low-dimensional kinetic parameter space ($\alpha$, $\beta$, $y_0$) rather than the full neural network parameter space.
\item A two-stage inference procedure combining maximum a posteriori (MAP) initialization with Hamiltonian Monte Carlo refinement is designed to support stable posterior inference and computationally efficient sampling.
\end{enumerate}

\section{Problem Formulation}
\begin{figure}[!t]
\centering
\includegraphics[width=0.7\linewidth]{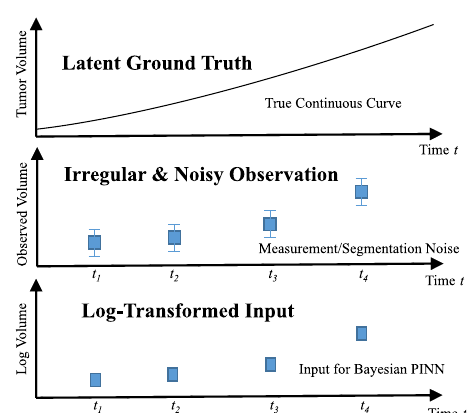}
\caption{Observation model with multiplicative noise and its log-transformed additive form. Tumor volumes are observed at discrete time points with measurement noise. Log transformation converts multiplicative noise into additive Gaussian noise, enabling likelihood-based modeling.}
\label{fig:observation}
\end{figure}

During clinical follow-up, patients typically undergo CT scans at several discrete time points, providing observations of tumor volume over time~\cite{nlst2011}. To model tumor growth dynamics and make predictions, we first define the observation model and introduce the Gompertz growth kinetics~\cite{norton1988}, laying the groundwork for the subsequent Bayesian PINN framework.

At discrete time points $\{t_i\}_{i=1}^{N_{\text{obs}}}$ the observed volumes satisfy
\begin{equation}
V_i^{\text{obs}} = V(t_i) \exp(\epsilon_i), \quad \epsilon_i \sim \mathcal{N}(0, \sigma_v^2),
\label{eq:obs_volume}
\end{equation}
where $V(t)$ denotes the true volume and $\epsilon_i$ represents observation noise.

For modeling convenience and to convert multiplicative noise into an additive form, we apply a log transformation to the observed volumes:
\begin{equation}
y(t) = \log V(t), \quad y_i^{\text{obs}} = \log V_i^{\text{obs}}.
\label{eq:log_transform}
\end{equation}

The observation model then becomes additive Gaussian noise:
\begin{equation}
y_i^{\text{obs}} = y(t_i) + \epsilon_i, \quad \epsilon_i \sim \mathcal{N}(0, \sigma_v^2),
\label{eq:obs_log}
\end{equation}
where $y(t)$ is the log-volume. This form corresponds to a Gaussian likelihood, which simplifies subsequent Bayesian modeling~\cite{yang2021}.

Tumor growth kinetics are described by the Gompertz model~\cite{norton1988}. In the volume domain,
\begin{equation}
\frac{dV(t)}{dt} = aV(t) - bV(t) \log V(t).
\label{eq:gompertz_volume}
\end{equation}

Letting $y(t) = \log V(t)$, the log-domain form is
\begin{equation}
\frac{dy(t)}{dt} = \alpha - \beta y(t).
\label{eq:gompertz_log}
\end{equation}

With the observation model and kinetic constraints defined above, we will next build a Bayesian PINN framework that combines physical constraints with data-driven learning for tumor growth modeling and uncertainty quantification.

\section{Bayesian Physics-Informed Neural Network}
\begin{figure}[!t]
\centering
\includegraphics[width=0.92\linewidth]{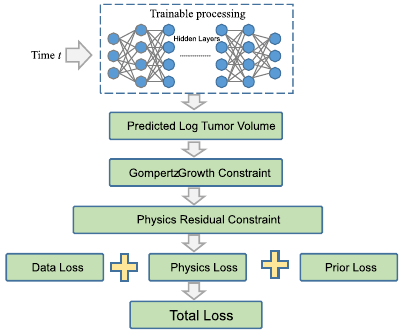}
\caption{Physics-informed neural network architecture for tumor growth modeling. The network maps time input to log-volume prediction while enforcing Gompertz dynamics via a physics residual term, combining data fitting and biological constraints.}
\label{fig:architecture}
\end{figure}

In this study, we construct a Physics Informed Neural Network (PINN)~\cite{raissi2019} to characterize tumor growth dynamics. The network takes time $t$ as input and outputs the predicted log tumor volume:
\begin{equation}
y(t) = \log V(t).
\label{eq:log_output}
\end{equation}

The neural network is denoted as
\begin{equation}
y_{\theta}(t) = f_{\theta}(t),
\label{eq:nn}
\end{equation}
where $\theta$ represents the network parameters. We use a multi-layer fully connected architecture and apply time normalization to improve training stability~\cite{wang2021a,wang2022}.

In the log domain, the Gompertz kinetics are given by~\cite{norton1988}
\begin{equation}
\frac{dy(t)}{dt} = \alpha - \beta y(t).
\label{eq:gompertz_kinetics}
\end{equation}

The physical residual is then defined as~\cite{wang2021a}
\begin{equation}
r(t) = \frac{dy_{\theta}(t)}{dt} - \bigl(\alpha - \beta y_{\theta}(t)\bigr).
\label{eq:residual}
\end{equation}

We impose the kinetic constraint at a set of discrete time points $\{t_j\}$, leading to the physics loss
\begin{equation}
\mathcal{L}_{\text{phys}} = \sum_j r(t_j)^2.
\label{eq:physics_loss}
\end{equation}

On top of the PINN framework, we introduce Bayesian modeling to capture parameter and prediction uncertainty~\cite{yang2021,neal1996}. The observation model is
\begin{equation}
y_i^{\text{obs}} = y_{\theta}(t_i) + \epsilon_i, \quad \epsilon_i \sim \mathcal{N}(0, \sigma_v^2),
\label{eq:bayesian_obs}
\end{equation}
with the corresponding likelihood
\begin{equation}
p(\mathbf{y}^{\text{obs}} \mid \theta, \alpha, \beta, \sigma_v) = \prod_{i=1}^{N_{\text{obs}}} \mathcal{N}\bigl(y_i^{\text{obs}} \mid y_{\theta}(t_i), \sigma_v^2\bigr).
\label{eq:likelihood}
\end{equation}

The prior distributions are set as follows: $\theta \sim \mathcal{N}(0, \sigma_w^2)$, $\alpha, \beta \sim \text{LogNormal}$ (to enforce positivity); $\sigma_v \sim \text{Half-Normal}$~\cite{polson2010}. This yields the posterior
\begin{equation}
p(\theta, \alpha, \beta, \sigma_v \mid \mathcal{D}) \propto p(\mathcal{D} \mid \theta, \alpha, \beta, \sigma_v) \cdot p(\theta) \cdot p(\alpha) \cdot p(\beta) \cdot p(\sigma_v).
\label{eq:posterior}
\end{equation}

For computational convenience in optimization and inference, we work with the negative logarithm, which gives an energy function composed of three terms~\cite{costa2024,zhu2018}:
\begin{equation}
\mathcal{L}_{\text{total}} = \mathcal{L}_{\text{data}} + \mathcal{L}_{\text{phys}} + \mathcal{L}_{\text{prior}}.
\label{eq:total_loss}
\end{equation}

Here $\mathcal{L}_{\text{data}}$ is the data term, $\mathcal{L}_{\text{phys}}$ the kinetic constraint term, and $\mathcal{L}_{\text{prior}}$ is
\begin{equation}
\mathcal{L}_{\text{prior}} = -\log p(\theta) - \log p(\alpha) - \log p(\beta) - \log p(\sigma_v).
\label{eq:prior_loss}
\end{equation}

\section{Inference and Uncertainty Quantification}
\begin{figure}[!t]
\centering
\includegraphics[width=\linewidth]{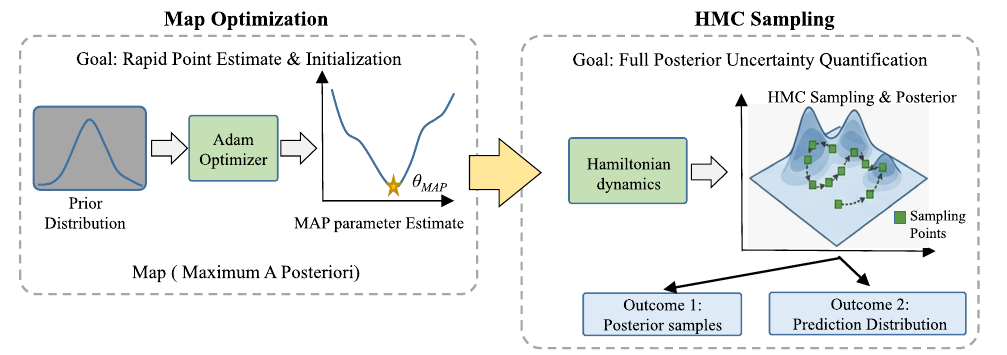}
\caption{Two-stage inference framework for Bayesian PINN. A MAP phase provides deterministic parameter initialization, followed by HMC sampling in a low-dimensional parameter space ($\alpha$, $\beta$, $y_0$) to obtain posterior distributions and uncertainty estimates.}
\label{fig:inference}
\end{figure}

After building the model, we first perform deterministic optimization via maximum a posteriori (MAP) estimation~\cite{yang2021}, which minimizes the negative log posterior:
\begin{equation}
\hat{\theta}, \hat{\alpha}, \hat{\beta}, \hat{\sigma}_v = \arg\min_{\theta, \alpha, \beta, \sigma_v} \mathcal{L}_{\text{total}}.
\label{eq:map}
\end{equation}

where the total energy function is $\mathcal{L}_{\text{total}} = \mathcal{L}_{\text{data}} + \mathcal{L}_{\text{phys}} + \mathcal{L}_{\text{prior}}$.

In implementation, we jointly optimize the network parameters $\theta$ and the kinetic parameters $\alpha, \beta$ using the Adam algorithm, and enforce positivity via a softplus transformation. The data term and the kinetic residual are evaluated at the observation points and at a set of collocation points, respectively.

Starting from the MAP solution, we then draw samples from the posterior using Hamiltonian Monte Carlo (HMC) to quantify uncertainty~\cite{betancourt2017,duane1987,chen2014,hoffman2014}. Sampling is performed only in the low-dimensional kinetic parameter space, which includes $\alpha$, $\beta$, and the initial condition $y_0$, thereby reducing computational complexity~\cite{hoffman2014,betancourt2017}.

HMC uses the negative log posterior (i.e., $\mathcal{L}_{\text{total}}$) to generate samples, with the Hamiltonian defined as
\begin{equation}
H = \mathcal{L}_{\text{total}} + K,
\label{eq:hamiltonian}
\end{equation}
where $K$ is the kinetic energy. The resulting set of posterior samples is denoted as $\{\alpha^{(s)}, \beta^{(s)}, y_0^{(s)}\}_{s=1}^{S}$.

Based on these posterior samples, we construct the posterior predictive distribution. For a given initial time $t_0$, the log-space prediction for sample $s$ is
\begin{equation}
y^{(s)}(t) = \frac{\alpha^{(s)}}{\beta^{(s)}} + \left(y_0^{(s)} - \frac{\alpha^{(s)}}{\beta^{(s)}}\right) \exp\bigl(-\beta^{(s)}(t - t_0)\bigr),
\label{eq:prediction_log}
\end{equation}
and the corresponding volume prediction is
\begin{equation}
V^{(s)}(t) = \exp\bigl(y^{(s)}(t)\bigr).
\label{eq:prediction_volume}
\end{equation}

The posterior predictive distribution is obtained by aggregating over all samples. We take the sample mean as the point prediction, and use the 2.5th and 97.5th percentiles to construct a 95\% credible interval.

Additionally, we compute the coverage probability of the prediction intervals at the observed time points, and analyze how the interval width evolves over future time windows to characterize the growth of uncertainty with the prediction horizon~\cite{costa2024}.

\section{Experimental Setup}
\begin{figure}[!htbp]
\centering
\includegraphics[width=0.92\columnwidth]{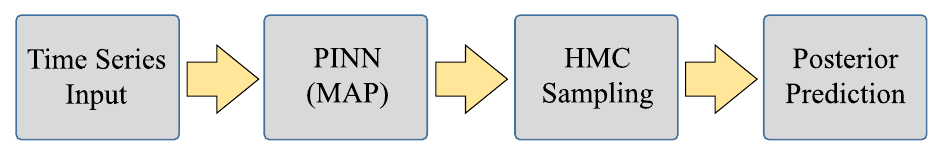}
\caption{Experimental pipeline for tumor volume extraction and modeling. The workflow includes NLST data selection, tumor segmentation, volume computation, and construction of longitudinal time series for Bayesian PINN training and evaluation.}
\label{fig:pipeline}
\end{figure}

This study uses longitudinal low-dose CT follow-up data from the National Lung Screening Trial (NLST)~\cite{nlst2011}. We selected 30 patients with at least three follow-up visits, and for each patient we took CT scans at three time points, resulting in 90 three-dimensional CT volumes. Tumor volumes were obtained by semi-automatic segmentation~\cite{ocana2024} and calculated using voxel spacing: $V_{\text{mm}^3} = N \cdot s_x \cdot s_y \cdot s_z$. The resulting volume sequences form longitudinal time series, which are log-transformed before being fed into the model. 

To evaluate predictive generalization under sparse longitudinal observations, a patient-wise temporal train--test split was adopted. For each patient, the first two CT observation time points were used for model training and posterior inference, while the final follow-up observation was strictly held out as unseen test data for evaluating predictive performance. This design avoids information leakage from future observations and better reflects the practical clinical setting of forecasting tumor progression from limited historical scans.

For model training, the PINN takes time $t$ as input and outputs the log volume $y(t) = \log V(t)$. The network has a fully connected architecture with three hidden layers of 64 units each and Tanh activation~\cite{wang2021a}. In the MAP stage, we jointly optimize the network parameters and the kinetic parameters using the Adam optimizer (learning rate $1\times10^{-3}$, 5000 epochs, random seed 42), and generate 200 collocation points within the observation interval. The model parameters are set to $\sigma_d = 0.2$, $\sigma_p = 0.5$, and each loss term is given a weight of 1. The kinetic parameters follow log-normal priors: $\log\alpha \sim \mathcal{N}(\log 0.2, 0.5^2)$, $\log\beta \sim \mathcal{N}(\log 0.05, 0.5^2)$~\cite{norton1988}.

In the HMC stage, to keep computation feasible, posterior inference is restricted to the low-dimensional physical parameter space ($\alpha$, $\beta$, $y_0$), thereby avoiding full Bayesian modeling over the high-dimensional neural network parameters. Sampling is performed using Hamiltonian Monte Carlo~\cite{betancourt2017,duane1987,chen2014,hoffman2014} with a step size of 0.01, 20 leapfrog steps, and a total of 400 samples (the first 100 used as burn-in). From the retained posterior samples, we compute predictions on a uniform time grid of 200 time points.

Model performance is evaluated by log-space and volume-space RMSE, 95\% credible interval coverage, relative CI width, and HMC acceptance rate. Uncertainty evolution is characterized by the ratio of CI width to posterior mean over future prediction windows~\cite{costa2024}.

\section{Results}

\subsection{Single-Patient Analysis} 
To evaluate model performance across different tumor growth patterns, we selected four representative cases corresponding to steady growth (Patient 100067), slow growth (Patient 100082), complex change (Patient 100153), and a challenging case (Patient 100166). For each case, we assess the model from three aspects: posterior fitting, extrapolation uncertainty, and posterior distribution of parameters~\cite{yang2021,costa2024}. 

\textbf{Case 1 (Patient 100067):}\par \begin{figure}[!t] \centering \includegraphics[width=0.8\linewidth]{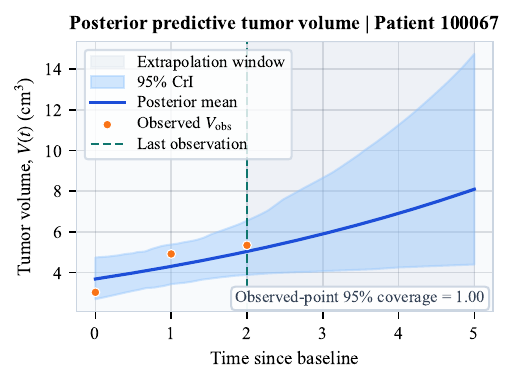} \caption{Posterior predictive tumor volume with observed data and 95\% credible interval (Patient 100067).} \label{fig:p100067_fit} \end{figure} \begin{figure}[!t] \centering \includegraphics[width=\linewidth]{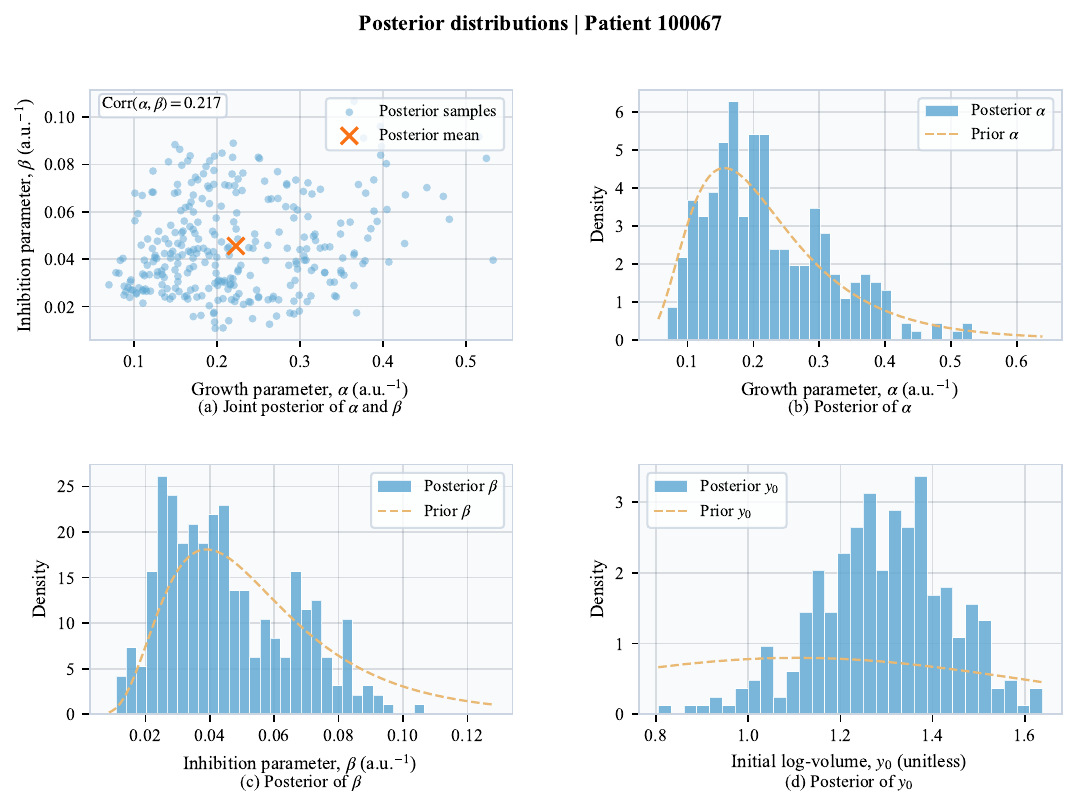} \caption{Posterior distributions of $\alpha$, $\beta$, and $y_0$ with joint structure (Patient 100067).} \label{fig:p100067_post} \end{figure} The model fits the observed data well (Fig.~\ref{fig:p100067_fit}); all observation points fall inside the 95\% credible interval, and uncertainty grows smoothly without abnormal divergence during extrapolation. The posterior distributions (Fig.~\ref{fig:p100067_post}) show a wide spread for $\alpha$, a relatively concentrated $\beta$, and a weak positive correlation between them.

\textbf{Case 2 (Patient 100082):}\par \begin{figure}[!t] \centering \includegraphics[width=0.8\linewidth]{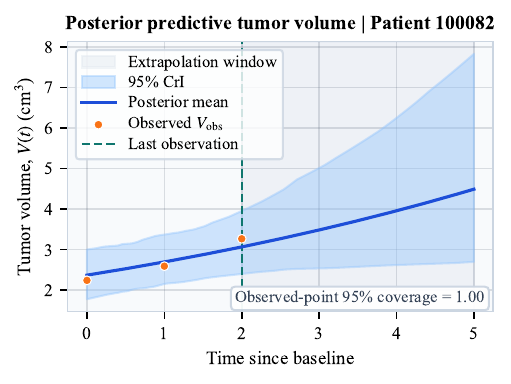} \caption{Posterior predictive tumor volume with observed data and 95\% credible interval (Patient 100082).} \label{fig:p100082_fit} \end{figure} \begin{figure}[!t] \centering \includegraphics[width=\linewidth]{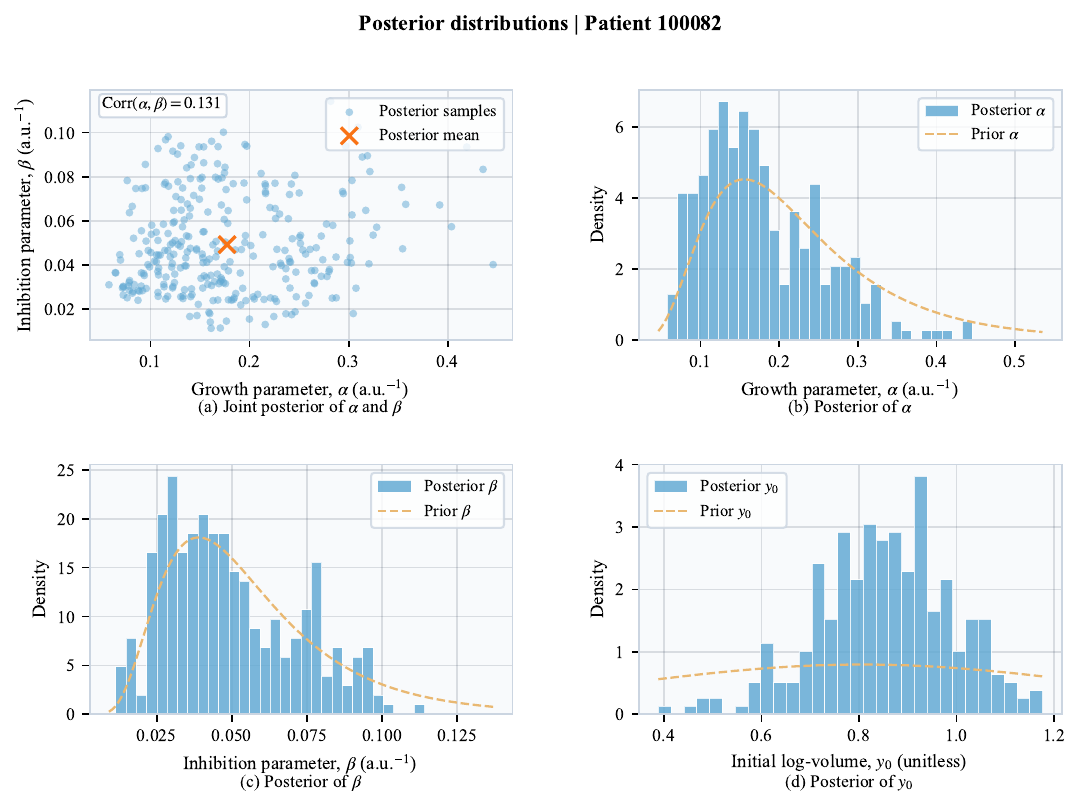} \caption{Posterior distributions of $\alpha$, $\beta$, and $y_0$ with joint structure (Patient 100082).} \label{fig:p100082_post} \end{figure} The fitted results align well with the observations (Fig.~\ref{fig:p100082_fit}); the prediction intervals are stable and expand smoothly into the extrapolation region. The posterior distributions (Fig.~\ref{fig:p100082_post}) indicate that both $\alpha$ and $\beta$ are fairly concentrated with weak correlation, suggesting good parameter identifiability and low uncertainty. 

\textbf{Case 3 (Patient 100153):}\par \begin{figure}[!t] \centering \includegraphics[width=0.8\linewidth]{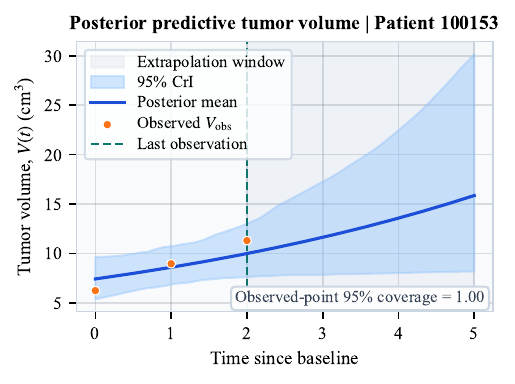} \caption{Posterior predictive tumor volume with observed data and 95\% credible interval (Patient 100153).} \label{fig:p100153_fit} \end{figure} \begin{figure}[!t] \centering \includegraphics[width=\linewidth]{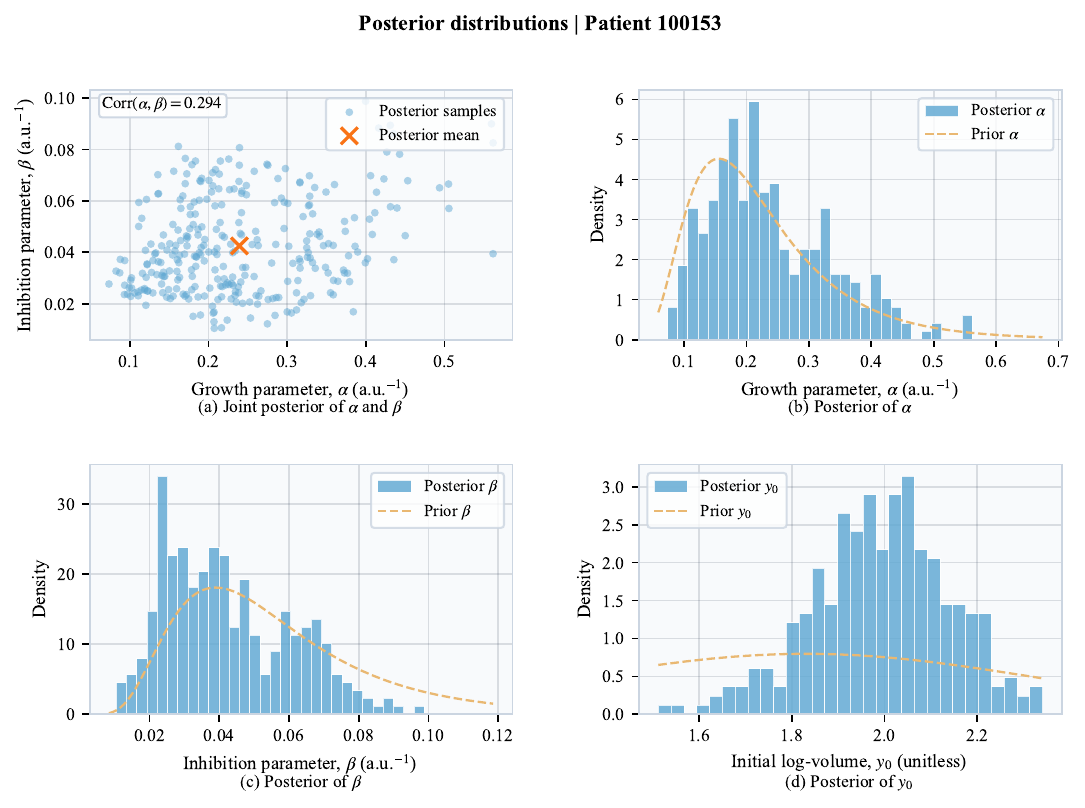} \caption{Posterior distributions of $\alpha$, $\beta$, and $y_0$ with joint structure (Patient 100153).} \label{fig:p100153_post} \end{figure} The model captures the rapid growth trend (Fig.~\ref{fig:p100153_fit}), and the prediction intervals widen noticeably in the extrapolation region. The posterior distributions (Fig.~\ref{fig:p100153_post}) show a wider spread for both $\alpha$ and $\beta$ along with a stronger correlation, reflecting increased parameter coupling and higher uncertainty. 

\textbf{Case 4 (Patient 100166):}\par \begin{figure}[!t] \centering \includegraphics[width=0.8\linewidth]{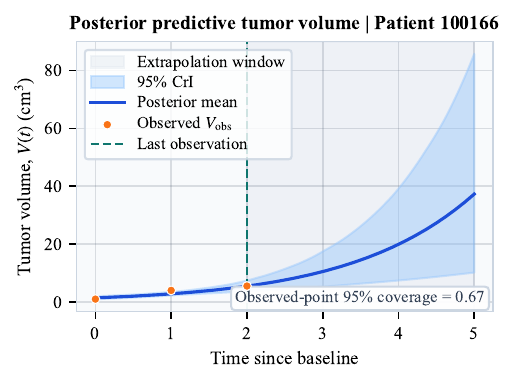} \caption{Posterior predictive tumor volume with observed data and 95\% credible interval (Patient 100166).} \label{fig:p100166_fit} \end{figure} \begin{figure}[!t] \centering \includegraphics[width=\linewidth]{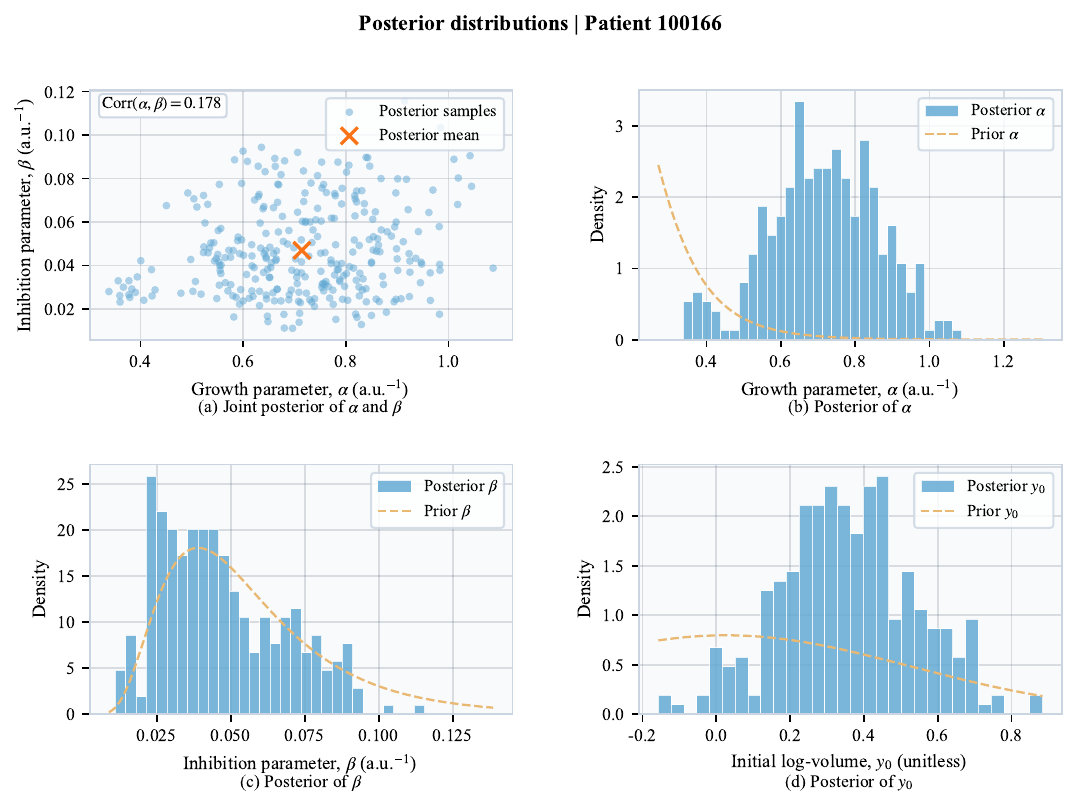} \caption{Posterior distributions of $\alpha$, $\beta$, and $y_0$ with joint structure (Patient 100166).} \label{fig:p100166_post} \end{figure} The fit deviates from the observed data (Fig.~\ref{fig:p100166_fit}), and the prediction intervals expand rapidly in the extrapolation region. The posterior distributions (Fig.~\ref{fig:p100166_post}) are highly dispersed and lack a stable structure, indicating poor parameter identifiability and unreliable predictions. These representative cases demonstrate that the model produces consistent uncertainty behavior across different tumor growth patterns: uncertainty increases gradually under complex dynamics, and predictive performance degrades markedly when parameter identifiability is poor.

\subsection{Multi-Patient Validation}

To assess generalization at the population level, we performed a cohort-wide analysis using longitudinal data from all 30 patients. Table~\ref{tab:cohort} summarizes the overall performance metrics, including log-space RMSE, relative credible interval width, and coverage. Table~\ref{tab:patients} provides parameter estimates and prediction results for the four representative patients (100067, 100082, 100153, 100166). Full patient-level results are provided as supplementary material (GitHub repository~\cite{github2026}).

\begin{table}[!t]
\centering
\caption{Cohort-level summary metrics ($n=30$).}
\label{tab:cohort}
\begin{tabular}{@{}lc@{}}
\toprule
\textbf{Metric} & \textbf{Value} \\
\midrule
RMSE (log) & $0.20 \pm 0.10$ \\
Rel.~CI width & $0.48$ \\
Coverage (95\%) & $0.95$ \\
\bottomrule
\end{tabular}
\end{table}

\begin{table*}[!t]
\centering
\caption{Representative patient-level results.}
\label{tab:patients}
\begin{tabular}{@{}ccccc@{}}
\toprule
\textbf{Patient ID} & \textbf{$\alpha$ (median)} & \textbf{$\beta$ (median)} & \textbf{RMSE (log)} & \textbf{Rel.~CI width} \\
\midrule
100067 & 0.201 [0.091, 0.417] & 0.042 [0.016, 0.086] & 0.251 & 0.500 \\
100082 & 0.161 [0.072, 0.341] & 0.045 [0.017, 0.094] & 0.155 & 0.476 \\
100153 & 0.217 [0.098, 0.447] & 0.039 [0.016, 0.080] & 0.244 & 0.478 \\
100166 & 0.715 [0.381, 0.983] & 0.043 [0.016, 0.089] & 0.732 & 0.666 \\
\bottomrule
\end{tabular}
\\[6pt]
\footnotesize Representative patient-level results (full results available in the supplementary repository).
\end{table*}

\begin{figure}[!t] \centering \subfloat[Population posterior.]{ \includegraphics[width=\linewidth]{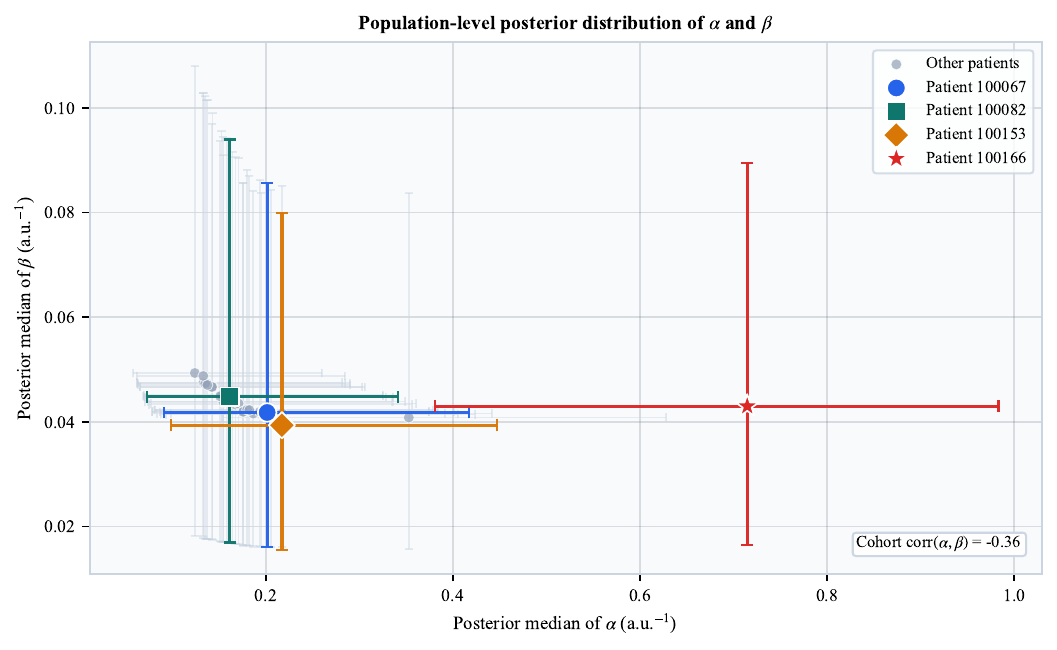} \label{fig:population_posterior} } \vspace{0.5em} \subfloat[Calibration curve.]{ \includegraphics[width=0.78\linewidth]{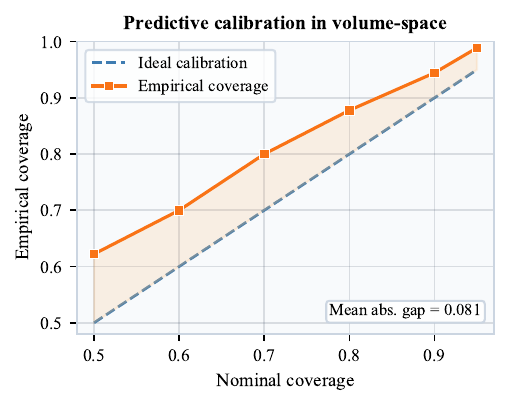} \label{fig:calibration_curve} } \caption{Population-level posterior distribution and predictive calibration of the Bayesian PINN framework.} \label{fig:population_calibration} \end{figure}

The cohort-level results (Table~I) indicate stable predictive performance (log-space RMSE: $0.20 \pm 0.10$) together with reliable uncertainty quantification (coverage: 0.95; relative credible interval width: 0.48). The parameters $\alpha$ and $\beta$ exhibit a consistent negative correlation at the population level (Fig.~\ref{fig:population_posterior})~\cite{liang2023}. In addition, the calibration curve (Fig.~\ref{fig:calibration_curve}) remains close to the ideal diagonal across all nominal coverage levels, with a mean absolute calibration gap of 0.081, indicating good uncertainty calibration.

\subsection{Comparison with Baseline Methods}

To evaluate predictive performance and uncertainty characterization under sparse and irregular longitudinal CT observations, the proposed framework was compared with six baseline methods, including Bayesian and deterministic variants of PINN-, Gompertz-, and Gaussian process-based models. All methods were trained on the same longitudinal observations and evaluated on held-out follow-up time points.

Table~\ref{tab:seven_methods} summarizes cohort-level prediction and uncertainty metrics across the seven methods. Among deterministic approaches, the Pure Gompertz model achieved relatively low prediction error (RMSE = 0.29318, MAE = 1.19552), reflecting the strong inductive bias of parametric kinetic models under relatively simple growth dynamics~\cite{swanson2001,byrne1995,norton1988}. However, deterministic approaches do not provide predictive uncertainty estimates.

The Pure PINN and Pure GP baselines showed higher prediction error and greater variability under sparse observations. In particular, the Pure PINN model produced the largest RMSE and MAE values, suggesting that purely data-driven neural modeling may become unstable in limited-data settings without explicit probabilistic inference or uncertainty regularization~\cite{wang2021a,wang2022}.

Among Bayesian approaches, the PINN + Bayesian variant produced relatively narrow credible intervals but showed substantial under-coverage, with a coverage deviation of approximately $-0.72$. This pattern suggests overconfident uncertainty estimates despite moderate prediction accuracy, consistent with previously reported limitations of approximate Bayesian deep learning methods~\cite{blundell2015,gal2016}. The Bayesian GP baseline improved uncertainty representation compared with deterministic models, although prediction error remained higher than that of the proposed framework.

The proposed Bayesian PINN framework achieved comparatively balanced performance across prediction accuracy and uncertainty calibration. Relative credible interval width and interval score remained stable without the severe under-coverage observed in the PINN + Bayesian variant.

\begin{figure}[!t]
\centering
\includegraphics[width=\linewidth]{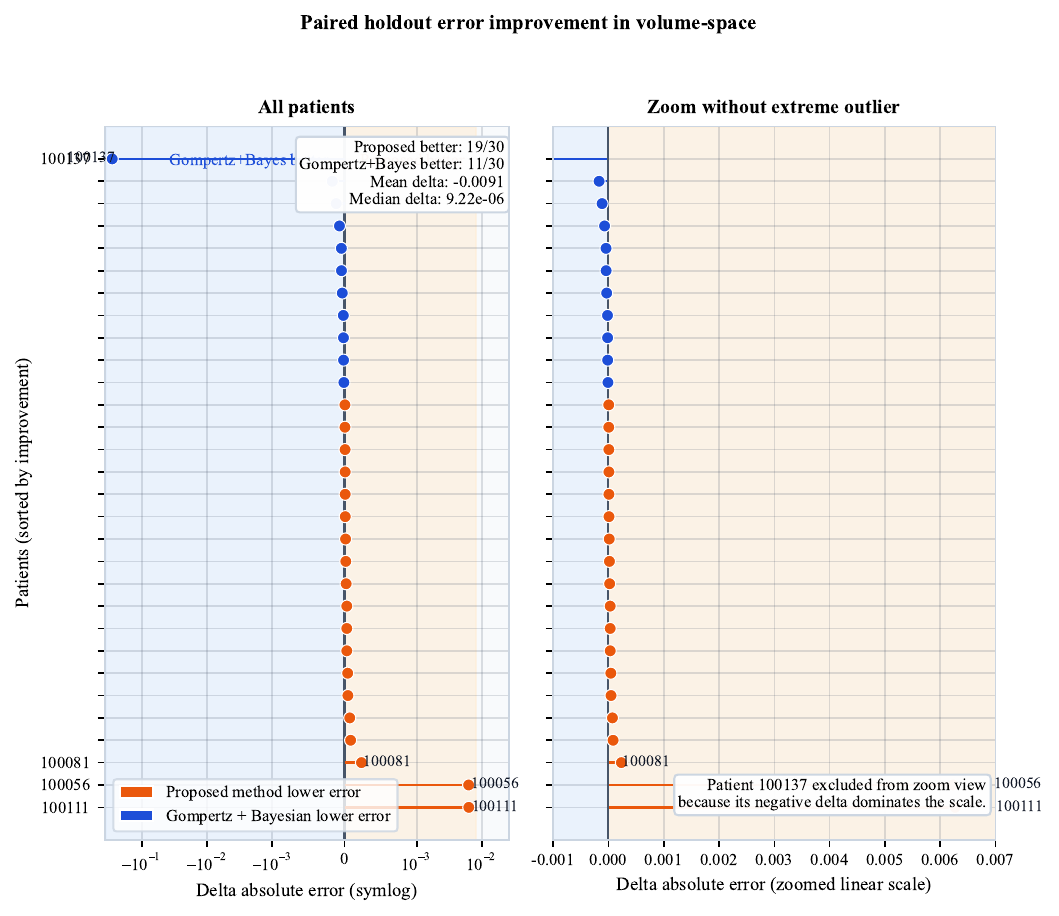}
\caption{Paired holdout error comparison between Bayesian PINN and Gompertz + Bayesian in volume-space. Positive values indicate lower error for the proposed framework. The right panel excludes one extreme outlier for visualization clarity.}
\label{fig:paired_improvement}
\end{figure}

The proposed framework and the Gompertz + Bayesian baseline produced similar cohort-level average errors, suggesting that averaged metrics alone may not fully reflect patient-level differences. To further examine this behavior, paired holdout error analysis was performed between the two methods (Fig.~\ref{fig:paired_improvement}).

As shown in Fig.~\ref{fig:paired_improvement}, the proposed Bayesian PINN achieved lower holdout error in 19 of 30 patients, whereas the Gompertz + Bayesian model performed better in 11 patients. Although the average improvement was modest, the results suggest more consistent patient-level refinement under heterogeneous growth patterns when the PINN component is incorporated. Performance differences were concentrated in a subset of patients with more complex dynamics.

Although the proposed Bayesian PINN achieves prediction accuracy comparable to existing Bayesian Gompertz models, it provides additional flexibility for representing heterogeneous tumor growth trajectories while maintaining biologically constrained uncertainty behavior. These results suggest that the primary advantage of the framework lies not only in point prediction accuracy, but also in its ability to provide uncertainty-aware longitudinal predictions under sparse clinical observations.

Overall, the comparative results highlight a trade-off between prediction accuracy, physical consistency, and uncertainty characterization under sparse longitudinal observations.

\begin{table*}[!t]
\centering
\caption{Comparison of prediction and uncertainty metrics across seven methods.}
\label{tab:seven_methods}
\setlength{\tabcolsep}{3pt}
\resizebox{\textwidth}{!}{%
\begin{tabular}{@{}lccccccccccc@{}}
\toprule
\textbf{Method} & \multicolumn{4}{c}{\textbf{RMSE (log)}} & \multicolumn{4}{c}{\textbf{MAE (volume)}} & \multicolumn{3}{c}{\textbf{Uncertainty (volume)}} \\
\cmidrule(lr){2-5}\cmidrule(lr){6-9}\cmidrule(l){10-12}
& \textbf{Value} & \textbf{$t$ p} & \textbf{W p} & \textbf{$d_z$} & \textbf{Value} & \textbf{$t$ p} & \textbf{W p} & \textbf{$d_z$} & \textbf{Rel. CI Width} & \textbf{Interval Score} & \textbf{Coverage Dev.} \\
\midrule
Bayesian PINN (proposed) & 0.19821 & N/A & N/A & N/A & 0.81756 & N/A & N/A & N/A & 0.78723 & 4.62139 & 0.01667 \\
PINN + Bayesian & 0.25686 & 0.0618 & 0.1142 & 0.35471 & 0.96947 & 0.2856 & 0.1519 & 0.19863 & 0.13327 & 27.72350 & -0.71667 \\
Pure PINN & 0.72243 & $1.69\times10^{-9}$ & $1.30\times10^{-8}$ & 1.57483 & 2.71723 & $2.14\times10^{-10}$ & $1.30\times10^{-8}$ & 1.73226 & N/A & N/A & N/A \\
Gompertz + Bayesian & 0.19814 & 0.1285 & 0.0010 & -0.28569 & 0.81822 & 0.4103 & 0.0577 & 0.15253 & 0.78769 & 4.62390 & 0.01667 \\
Bayesian GP & 0.52563 & 0.0024 & $9.98\times10^{-7}$ & 0.60747 & 1.90945 & $1.48\times10^{-5}$ & $5.14\times10^{-6}$ & 0.94820 & 0.52004 & 28.61126 & -0.71667 \\
Pure GP & 0.39323 & 0.2172 & 0.0197 & 0.23029 & 1.08203 & 0.1474 & 0.0577 & 0.27179 & N/A & N/A & N/A \\
Pure Gompertz & 0.29318 & 0.1621 & 0.1191 & 0.26191 & 1.19552 & 0.2440 & 0.2129 & 0.21710 & N/A & N/A & N/A \\
\bottomrule
\end{tabular}%
}

\vspace{2pt}
\footnotesize
\textit{Note:} Paired statistical tests use the proposed Bayesian PINN as the reference. W p denotes the Wilcoxon signed-rank test p-value. For RMSE, tests are performed on patient-level squared log errors; for MAE, tests are performed on patient-level absolute volume errors. Lower RMSE/MAE and interval scores indicate better performance; coverage deviation is empirical 95\% coverage minus 0.95.

\end{table*}

\section{Discussion}

The Bayesian PINN framework enables uncertainty-aware longitudinal prediction through posterior predictive credible intervals and calibrated uncertainty metrics. In most patients, the model produces stable and reasonably calibrated prediction intervals; under sparse observations or more complex growth patterns, interval widening reflects increased predictive uncertainty~\cite{costa2024}. In practice, uncertainty-aware prediction may help distinguish between relatively stable and potentially high-risk tumor progression patterns when only limited longitudinal follow-up scans are available.

Methodologically, the framework separates neural network training from low-dimensional Bayesian inference~\cite{yang2021}. Posterior inference is restricted to a low-dimensional kinetic parameter space, improving computational feasibility and parameter interpretability. Compared with deterministic PINN models, the Bayesian PINN provides more stable uncertainty estimates under sparse observations. Compared with approximate Bayesian approaches~\cite{kingma2014}, HMC can better preserve posterior structure under sparse observations~\cite{betancourt2017,duane1987}. The combined MAP and HMC strategy provides a practical balance between inference quality and computational cost~\cite{yang2021,betancourt2017}.

The goal of this study is not to introduce a fundamentally new Bayesian inference algorithm, but to develop a practically usable framework for uncertainty-aware tumor growth prediction under sparse longitudinal observations. The main advantage of the Bayesian PINN lies not solely in lower average prediction error, but in combining calibrated uncertainty with more consistent patient-level refinement relative to the Bayesian Gompertz baseline. 

Limitations include the modest cohort size (30 patients), sparse longitudinal observations, the simplified kinetic formulation, and the computational cost of HMC sampling. Because each patient contains only three CT follow-up time points, the statistical comparisons should be interpreted as exploratory rather than definitive clinical evidence. These factors limit the evaluation of long-term predictive performance and more complex biological mechanisms. Future work will focus on larger longitudinal cohorts, multi-modal clinical data, more flexible kinetic models, and computationally efficient Bayesian inference strategies.

\section{Conclusion}

To address the clinical challenge of sparse, irregularly sampled longitudinal CT data with observational uncertainty, we have proposed a Bayesian PINN framework that integrates Gompertz kinetics with Bayesian inference, unifying tumor growth modeling and uncertainty quantification~\cite{yang2021,norton1988}.

The model captures heterogeneous tumor growth patterns while maintaining stable predictive behavior across the cohort. It provides well-calibrated prediction intervals while maintaining reasonable point prediction accuracy, outperforming the deterministic 
PINN and addressing the lack of uncertainty quantification in traditional parametric models~\cite{raissi2019}.

Under limited longitudinal follow-up data, the Bayesian PINN provides biologically constrained prediction with calibrated uncertainty estimates. These results suggest that combining physics-constrained modeling with low-dimensional Bayesian inference may support uncertainty-aware tumor growth assessment under sparse longitudinal follow-up observations~\cite{raissi2019,yang2021}.

Future work will focus on multi-modal data fusion, more flexible kinetic models, and efficient Bayesian inference methods to further improve model performance and clinical scalability.

\noindent\textbf{Ethics Statement:}
The NLST dataset used in this study is publicly available and de-identified. The imaging data were accessed through the Cancer Imaging Archive (TCIA) in accordance with the NCI data use agreement. The original NLST study was conducted with institutional review board approval and informed consent from participants~\cite{nlst2011}. This retrospective analysis used only de-identified imaging data.

\section*{Acknowledgments}
\addcontentsline{toc}{section}{Acknowledgments}

This work was supported by the Department of Mathematics at Vanderbilt University. The author also acknowledges the guidance and support from Prof. Glenn F. Webb through an independent study.

\section*{Data Availability}
\addcontentsline{toc}{section}{Data Availability}

The code and data for Bayesian PINN tumor modeling are available at GitHub~\cite{github2026}.

\bibliographystyle{IEEEtran}
\bibliography{references}

@article{raissi2019,
  author = {Raissi, M. and Perdikaris, P. and Karniadakis, G. E.},
  title = {{Physics-Informed Neural Networks}: a deep learning framework for solving forward and inverse problems involving nonlinear partial differential equations},
  journal = {Journal of Computational Physics},
  volume = {378},
  pages = {686--707},
  year = {2019}
}

@article{yang2021,
  author = {Yang, X. and Meng, X. and Karniadakis, G. E.},
  title = {{B-PINNs}: Bayesian physics-informed neural networks for forward and inverse problems with noisy data},
  journal = {Journal of Computational Physics},
  volume = {425},
  pages = {109913},
  year = {2021}
}

@article{rackauckas2020,
  author = {Rackauckas, C. and Nie, Q.},
  title = {Universal differential equations for scientific machine learning},
  journal = {Proceedings of the Royal Society A},
  volume = {476},
  number = {2242},
  pages = {20200237},
  year = {2020}
}

@article{costa2024,
  author = {Costa, E. A. and others},
  title = {Physics-informed neural network uncertainty assessment through Bayesian inference},
  journal = {IFAC-PapersOnLine},
  volume = {58},
  number = {14},
  pages = {652--657},
  year = {2024}
}

@article{nlst2011,
  author = {{National Lung Screening Trial Research Team}},
  title = {Reduced lung-cancer mortality with low-dose computed tomographic screening},
  journal = {New England Journal of Medicine},
  volume = {365},
  number = {5},
  pages = {395--409},
  year = {2011}
}

@article{liu2025,
  author = {Liu, L. and Wang, Y. and Xu, Q. and Xu, X.},
  title = {Data-driven parameter identification for tumor growth models},
  journal = {arXiv},
  note = {preprint arXiv:2511.15940},
  year = {2025}
}

@article{liang2023,
  author = {Liang, B. and others},
  title = {Bayesian inference of tissue heterogeneity for individualized prediction of glioma growth},
  journal = {IEEE Transactions on Medical Imaging},
  volume = {42},
  number = {10},
  pages = {2865--2875},
  year = {2023}
}

@article{rodrigues2024,
  author = {Rodrigues, J. A.},
  title = {Using physics-informed neural networks ({PINNs}) for tumor cell growth modeling},
  journal = {Mathematics},
  volume = {12},
  number = {8},
  pages = {1195},
  year = {2024}
}

@article{lorenzo2024,
  author = {Lorenzo, G. and others},
  title = {Patient-specific, mechanistic models of tumor growth incorporating artificial intelligence and big data},
  journal = {Annual Review of Biomedical Engineering},
  volume = {26},
  pages = {529--560},
  year = {2024}
}

@article{swanson2001,
  author = {Swanson, K. R. and Alvord Jr., E. C. and Murray, J. D.},
  title = {A quantitative model for differential motility of gliomas},
  journal = {Mathematical and Computer Modelling},
  volume = {33},
  pages = {663--674},
  year = {2001}
}

@article{byrne1995,
  author = {Byrne, H. and Chaplain, M. A. J.},
  title = {Growth of nonnecrotic tumors in the presence and absence of inhibitors},
  journal = {Mathematical Biosciences},
  volume = {130},
  pages = {151--165},
  year = {1995}
}

@article{norton1988,
  author = {Norton, L.},
  title = {A Gompertzian model of human breast cancer growth},
  journal = {Cancer Research},
  volume = {48},
  number = {24},
  pages = {7067--7071},
  year = {1988}
}

@article{zhu2018,
  author = {Zhu, Y. and Zabaras, N. and Perdikaris, P. and Karniadakis, G. E.},
  title = {Bayesian deep convolutional encoder--decoder networks for surrogate modeling},
  journal = {Journal of Computational Physics},
  volume = {366},
  pages = {415--447},
  year = {2018}
}

@article{betancourt2017,
  author = {Betancourt, M.},
  title = {A conceptual introduction to {Hamiltonian Monte Carlo}},
  journal = {arXiv},
  note = {preprint arXiv:1701.02434},
  year = {2017}
}

@article{duane1987,
  author = {Duane, S. and Kennedy, A. D. and Pendleton, B. J. and Roweth, D.},
  title = {{Hybrid Monte Carlo}},
  journal = {Physics Letters B},
  volume = {195},
  number = {2},
  pages = {216--222},
  year = {1987}
}

@inproceedings{chen2014,
  author = {Chen, T. and Fox, E. B. and Guestrin, C.},
  title = {Stochastic gradient {Hamiltonian Monte Carlo}},
  booktitle = {International Conference on Machine Learning (ICML)},
  pages = {1683--1691},
  year = {2014}
}

@article{hoffman2014,
  author = {Hoffman, M. D. and Gelman, A.},
  title = {The {No-U-Turn} sampler: adaptively setting path lengths in {Hamiltonian Monte Carlo}},
  journal = {Journal of Machine Learning Research},
  volume = {15},
  pages = {1593--1623},
  year = {2014}
}

@article{polson2010,
  author = {Polson, N. G. and Scott, J. G.},
  title = {Shrink globally, act locally: sparse {Bayesian} regularization and prediction},
  journal = {Journal of the Royal Statistical Society B},
  volume = {72},
  number = {4},
  pages = {465--483},
  year = {2010}
}

@inproceedings{kingma2014,
  author = {Kingma, D. P. and Welling, M.},
  title = {Auto-encoding variational Bayes},
  booktitle = {International Conference on Learning Representations (ICLR)},
  address = {Banff, AB, Canada},
  year = {2014}
}

@book{neal1996,
  author = {Neal, R. M.},
  title = {Bayesian Learning for Neural Networks},
  address = {New York, NY, USA},
  publisher = {Springer},
  year = {1996}
}

@article{wang2021a,
  author = {Wang, S. and Teng, Y. and Perdikaris, P.},
  title = {Understanding and mitigating gradient pathologies in physics-informed neural networks},
  journal = {SIAM Journal on Scientific Computing},
  volume = {43},
  number = {5},
  pages = {A3055--A3081},
  year = {2021}
}

@article{wang2022,
  author = {Wang, S. and Yu, X. and Perdikaris, P.},
  title = {When and why {PINNs} fail to train: a neural tangent kernel perspective},
  journal = {Journal of Computational Physics},
  volume = {449},
  pages = {110768},
  year = {2022}
}

@article{deisenroth2015,
  author = {Deisenroth, M. P. and Fox, D. and Rasmussen, C. E.},
  title = {Gaussian processes for data-efficient learning in robotics and control},
  journal = {IEEE Transactions on Pattern Analysis and Machine Intelligence},
  volume = {37},
  number = {2},
  pages = {408--423},
  year = {2015}
}

@article{nickisch2008,
  author = {Nickisch, H. and Rasmussen, C. E.},
  title = {Approximations for binary Gaussian process classification},
  journal = {Journal of Machine Learning Research},
  volume = {9},
  pages = {2035--2078},
  year = {2008}
}

@inproceedings{blundell2015,
  author = {Blundell, C. and others},
  title = {Weight uncertainty in neural networks},
  booktitle = {International Conference on Machine Learning (ICML)},
  pages = {1613--1622},
  year = {2015}
}

@inproceedings{gal2016,
  author = {Gal, Y. and Ghahramani, Z.},
  title = {Dropout as a {Bayesian} approximation: representing model uncertainty in deep learning},
  booktitle = {International Conference on Machine Learning (ICML)},
  pages = {1050--1059},
  year = {2016}
}

@article{ocana2024,
  author = {Ocana-Tienda, B. and others},
  title = {Growth dynamics of lung nodules: implications for classification in lung cancer screening},
  journal = {Cancer Imaging},
  volume = {24},
  number = {1},
  pages = {113},
  year = {2024}
}

@article{mishra2022,
  author = {Mishra, S. and Molinaro, R.},
  title = {Estimates on the generalization error of physics-informed neural networks for inverse problems of {PDEs}},
  journal = {IMA Journal of Numerical Analysis},
  volume = {42},
  number = {2},
  pages = {981--1022},
  year = {2022}
}

@article{kisbenedek2026,
  author = {Kisbenedek, L. and Kovacs, L. and Drexler, D. A.},
  title = {Physics-informed neural network for parameter inference in a tumor model},
  journal = {Mathematics},
  volume = {14},
  number = {7},
  pages = {1102},
  year = {2026}
}

@misc{github2026,
  author = {Kong, L.},
  title = {Code and manuscript-facing results for Bayesian PINN lung tumor growth modeling},
  year = {2026},
  howpublished = {\url{https://github.com/Lorettakong/bayesian-pinn-lung-tumor-growth}},
  note = {GitHub repository, accessed May 9, 2026}
}

\end{document}